\documentclass[10pt,twocolumn,letterpaper]{article}

\usepackage{iccv}
\usepackage{times}
\usepackage{epsfig}
\usepackage{graphicx}
\usepackage{amsmath}
\usepackage{amssymb}
\usepackage{rotating}
\usepackage{multirow}
\usepackage[pagebackref=true,breaklinks=true,letterpaper=true,colorlinks,bookmarks=false]{hyperref}

\iccvfinalcopy

\begin{document}

\title{Examining the Impact of Blur on Recognition by Convolutional Networks}

\author{Igor Vasiljevic\\
University of Chicago\\
{\tt\small ivasiljevic@uchicago.edu}
\and
Ayan Chakrabarti\\
TTI-Chicago\\
{\tt\small ayanc@ttic.edu}
\and
Gregory Shakhnarovich\\
TTI-Chicago\\
{\tt\small greg@ttic.edu}
}

\maketitle

\begin{abstract}
  State-of-the-art algorithms for many semantic visual tasks are based on the use of convolutional neural networks. These networks are commonly trained, and evaluated, on large annotated datasets of artifact-free high-quality images. In this paper, we investigate the effect of one such artifact that is quite common in natural capture settings: optical blur. We show that standard network models, trained only on high-quality images, suffer a significant degradation in performance when applied to those degraded by blur due to defocus, or subject or camera motion. We investigate the extent to which this degradation is due to the mismatch between training and input image statistics. Specifically, we find that fine-tuning a pre-trained model with blurred images added to the training set allows it to regain much of the lost accuracy.  We also show that there is a fair amount of generalization between different degrees and types of blur, which implies that a single network model can be used robustly for recognition when the nature of the blur in the input is unknown. We find that this robustness arises as a result of these models learning to generate blur invariant representations in their hidden layers. Our findings provide useful insights towards developing vision systems that can perform reliably on real world images affected by blur.
\end{abstract}

\section{Introduction}
\label{sec:intro}

Recent years have seen tremendous progress in the development of computer vision algorithms for semantic tasks such as image classification~\cite{resnet, alexnet,vgg16}, object detection~\cite{rcnn,frcnn}, and semantic segmentation~\cite{deeplab,zoomout}, with most modern state-of-the-art methods based on convolutional neural networks. These methods owe much of their success to the availability of large image datasets~\cite{imagenet,voc} with ground truth data collected through human annotation. To build datasets at such scales, researchers have had to rely on images freely shared by regular users. But there is significant potential for a disparity between the low-level statistics of images a vision algorithm receives as training input, and publicly shared photographs. This is because users tend to upload photographs, often selected from among multiple trials of capturing the same object, that are high-quality and free of artifacts---\eg saturation, distortions, and motion and defocus blur. This is a concern since modern methods are not only trained, \emph{but also evaluated}, on these datasets.

In this paper, we concentrate on blur, since it is likely to affect images taken even with a high-quality camera, especially in a setting where the user is not concerned about image quality (\eg, when the image is being taken for a vision application, rather than for photography), or when the image is being captured automatically by a device or robot. We are also motivated by the recent findings showing that modern image classification methods exhibit significantly lower accuracies when evaluated on blurred images~\cite{arizona,richardwebster2016psyphy,oh2016faceless,ding2016trunk}.

\begin{figure*}[!t]
  \centering
  \includegraphics[width=0.88\textwidth]{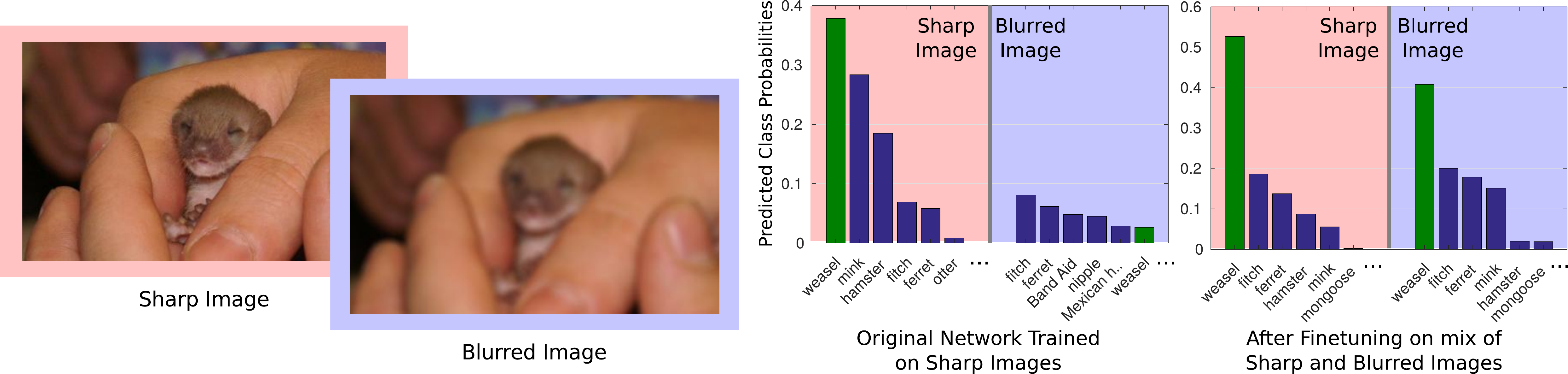}
  \caption{State-of-the-art image classification networks like VGG-16~\cite{vgg16} perform poorly on blurred input (left), when using model weights trained on high-quality sharp image datasets (center). However, while they often make erroneous predictions in terms of the most likely classes for a blurred image, they do so with lower confidence---producing distributions that are higher-entropy than those for sharp images. However, this drop in performance is largely an artifact of being trained without any blurred examples. We find that by fine-tuning the model on a mix of blurred and sharp images for just a few epochs, allows it to perform well on both sharp and blurred inputs (right).}
  \label{fig:teaser}
\end{figure*}

Our first contribution is a systematic examination of the effect of blur on the performance of convolutional neural networks for image recognition. We confirm that introduction of even a moderate blur hurts performance of networks trained on sharp images, and find that the resulting unreliable predictions are accompanied by underlying high-entropy class distributions (\eg, see Fig.~\ref{fig:teaser}). This can allow vision systems using such networks to account for the low confidence of their predictions (and, say, ask the user to take another image).

Our second contribution is the insight we offer into the mechanism by which blur causes such degradation, and a practical recipe for combating it. We demonstrate that much of performance drop under blur is a consequence of models being trained only on sharp images, and not due to an intrinsic absence of information in blurry images, or to a deficiency in the networks' architectures. By fine-tuning with a mix of blurry and sharp images for only three epochs, we find that a network trained only on sharp images is able to recover most of its lost accuracy on blurry images.  Interestingly, a model fine-tuned with a fairly diverse range of blurs performs almost on par with a network trained and evaluated on a single blur type (which in-turn performs worse on other blurs).  This demonstrates the ability of the network to learn a level of blur \emph{invariance}.

Our third contribution is our finding that fine-tuning for blur partially generalizes across different types of blur---networks fine-tuned on images blurred with complex camera shake kernels significantly improve performance on defocus and subject motion-blurred images.  Finally, in addition to image classification, we find that these observations also largely carry over to the task of semantic segmentation, where fine-tuning with blur is able to improve accuracy in both identifying and localizing objects in blurred images.

Broadly, our findings reinforce the fact that convolutional neural networks are resilient---when presented with an out-of-distribution input, they are able to signal a low-confidence in their predictions, and are quickly able to adapt when provided with additional training data, without any modification in architecture. Moreover, our experiments provide insight for designing vision systems that need to make reliable predictions in a practical, non-idealized, setting, by reasoning with images that may be potentially degraded by an unknown blur kernel.

\section{Background \& Related Work}
\label{sec:rw}

Blur is a degradation in image quality caused by camera sensor pixels averaging light from overlapping regions in the scene, typically due to defocus or motion in the capture interval. Often, an observed blurry image can be well-modeled as a convolution of a \emph{latent} sharp image (\ie, the image that would have been captured in the absence of blur), and a blur kernel or point-spread-function \footnote{The effective blur kernel can sometimes vary spatially across the image, \eg, due to depth variations for defocus blur, or rotational or out-of-plane subject and camera motion blur. In this work, we concentrate only on spatially uniform blur for ease of analysis, but it may be appropriate to expect a mix of blur kernels acting on an image in some settings.}. In the case of defocus blur, this kernel is an image of the lens aperture (typically a disk) scaled by a factor proportional to the distance of the imaged object from the focal plane. For motion blur, it is the projection of the moving object or camera's trajectory during the exposure interval. Both kinds of blur kernels are ``low-pass'' filters, \ie, they lead to a loss or attenuation of high-frequency image detail.  Camera shake kernels exhibit more variability than defocus blur, and typically correspond to arbitrary 2D motion trajectories ~\cite{levin,ayan}.

Reversing the effect of blur to obtain a sharp image from a blurred observation is an ill-posed inverse problem, especially when the blur is unknown. Graphics and vision researchers have made significant progress on this problem across the last decade~\cite{fergus06,levin,tomer,zoran}---including with recent neural network-based methods~\cite{ayan,schuler,hirsch,liu}. Nevertheless, image deblurring remains a challenging and computationally expensive task. However, our goal in this paper is different. Instead of processing blurred images to recover sharp photographs for human consumption, we seek to understand, and ameliorate, the effect of using these images as input to algorithms for recognition.

Nearly all state-of-the-art computer vision algorithms for semantic visual tasks rely on the use of convolutional neural networks~\cite{deeplab,rcnn,resnet,alexnet,zoomout,frcnn,vgg16}. Critical to their success is the ability to train on large annotated datasets, like Imagenet~\cite{imagenet} and Pascal VOC~\cite{voc}---both of which contain photographs downloaded from the photo sharing website \url{flickr.com}. Therefore, these datasets contain images that users---amateurs and professional photographers alike---have \emph{chosen} to upload, and are consequently of high-quality with few artifacts.

Since standard recognition benchmarks perform their evaluation on held-out portions of the same dataset, the reported performance of state-of-the-art algorithms can at best be interpreted to accurately characterize their expected accuracy on similar high-quality image data. Evaluation of standard neural network-based methods for image classification on images degraded by Gaussian blur in~\cite{arizona,richardwebster2016psyphy} shows a significant drop in classification accuracy due to blur. In concurrent work~\cite{zhou}, Zhou \etal also discuss loss in accuracy caused by various image degradations (including blur), and include preliminary experiments that suggest that this can be overcome to some extent by fine-tuning the initial layers (of Alexnet~\cite{alexnet}) on degraded data. Recent work has also considered on the effect of blur on networks trained for face recognition~\cite{faceblur,oh2016faceless,ding2016trunk}, while Ullman \etal~\cite{mirc} contrasted the drop in the accuracy of computational recognition to that of humans. Finally, Szegedy \etal~\cite{intrigue} found that carefully optimized small-magnitude perturbations could cause network models to produce erroneous estimates. In contrast, our focus is on errors due to a naturally prevalent form of image degradation---blur.

In this paper, we conduct a more thorough examination of the degradation caused by blur to the performance of modern convolutional neural network models, and the extent to which this degradation can be remedied and network models made robust in the presence of variability in the degree and type of unknown blur in an input image. We find, perhaps surprisingly, that there is a fair bit of generalization across certain blur types---training with moderate blur improves performance on severely blurred images, training with either defocus and camera motion blur kernels improves performance on the other, but that there is limited generalization from radially symmetric defocus blur to oriented one-dimensional motion kernels.

We show that while precise knowledge of the blur in an image is helpful (we see a slight improvement when using a model trained with only that blur kernel), recognition with unknown blur can be made almost as robust by using models trained with a diverse set of kernels. This is surprising since in the context of deblurring, blind deconvolution is a much harder problem than deconvolution with a known kernel. We find that models trained with different blurs achieve robustness to variability in blur by computing blur-invariant features representations in its internal hidden layers. We also demonstrate that even when the kernel is known, deblurring the input image using state of the art method and then using the original classifier trained on sharp images is in fact inferior to the mixed-tuning regime we propose.

\begin{figure}[!t]
  \label{fig:blurs}
  \centering
  \includegraphics[width=17em]{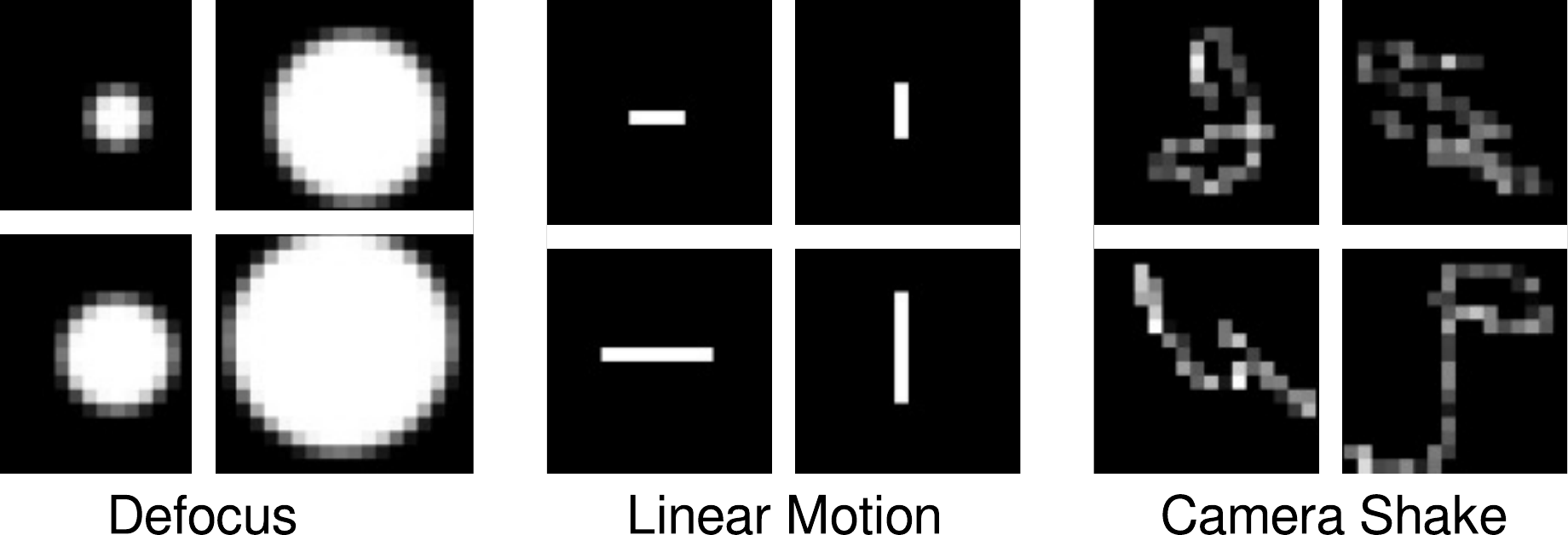}%
  \caption{Example blur kernels used in our experiments. We use disk kernels to simulate defocus blur, oriented box kernels to simulate uniform subject motion, as well as arbitrary trajectories (with varying intensities) for camera shake.}
\end{figure}

Therefore, we find that convolutional neural networks can succeed despite degradations in image quality from blur. This is consistent with such networks having been used quite successfully for the classification of very low-quality images, albeit on smaller benchmarks---\eg, CIFAR-100~\cite{cifar} that contains extremely low-resolution images of size $32\times 32$. In this context, recently Peng \etal~\cite{kate} explored the potential of jointly training with high-resolution images to boost performance on low-resolution inputs. Our work also demonstrates the success of neural networks in the face of lower image quality, with joint training with images of varying quality, to deal with degradation due to optical blur for large-scale visual recognition.

\section{Imagenet Classification on Blurred Images using Pre-trained Network Models}
\label{sec:orig}

We begin by evaluating the performance of a popular convolutional neural network model---VGG-16\footnote{The supplementary material includes results on ResNet~\cite{resnet}, which exhibits similar performance degradation with blur.}~\cite{vgg16}---on blurred versions of the Imagenet~\cite{imagenet} standard 2012 validation set. This model was trained on the Imagenet training set that comprised of largely sharp, high-quality images. We generate the blurred versions of the validation set by convolution with a range of different blur kernels.

Note that \cite{vgg16} resize their input to one or more pre-determined scales (to match their smaller side to the scale value), and then compute class probability distributions by feeding this resized image to their network model. Since Imagenet contains images of various sizes, rather than blur the original images directly, we first resize the image to a fixed scale (384 in our experiments) and then convolve the result with the blur kernel. This ensures that the effective blur for a chosen kernel is consistent across images, irrespective of their original size. We quantize the blurred intensities to form 8-bit images, and feed these into the classification networks at different scales.

We report the performance VGG-16 network~\cite{vgg16} for a diverse set of blur kernels (see Fig.~\ref{fig:blurs}) in Table~\ref{tab:origAll}. We consider disk kernels of different radii $r$ (that have a constant value within their radius, and zero outside) to simulate defocus blur, and with horizontal and vertical box kernels of single pixel width and different lengths that correspond to motion blur (for uniform linear motion). To simulate camera shake blur, we use the code from \cite{ayan} to generate a set of 100 random kernels as splines representing arbitrary motion trajectories, with intensity variation along the trajectory. We generate these splines using a set of six control points on a $15\times 15$ grid, and center these to yield $17\times 17$ blur kernels. We blur each image in the validation set with a different kernel---chosen deterministically across different test settings. For completeness, we also report results on blurring with Gaussian kernels of different standard deviations $\sigma$ considered in \cite{arizona}. All kernels are normalized to be unit sum, \ie, they have a ``DC value'' of one.

\begin{table}[!t]
  \centering
  {\footnotesize
    ~~~~~~~~~~~~~~~~~~~{\bf Scales}\vspace{0.5em}\\
    \parbox[c]{1.1em}{\rotatebox{90}{\bf Blur Type}}\parbox[c]{30em}{
    \begin{tabular}{|c||c|c|c|c|}
      \hline
      &128 & 256 & 256+512 & 512\\\hline\hline
      Sharp & 76.07\% & 90.88\% & 92.17\% & 90.76\%\\\hline
      \hspace{-0.7em}Defocus r=2(D2)\hspace{-0.7em}~& 74.83\% & 88.06\% & 89.00\% & 85.43\%\\
      \hspace{-0.7em}Defocus r=4(D4)\hspace{-0.7em}~& 68.48\% & 81.48\% & 80.93\% & 66.86\%\\
      \hspace{-0.7em}Defocus r=6(D6)\hspace{-0.7em}~& 61.03\% & 72.69\% & 67.86\% & 40.64\%\\
      \hspace{-0.7em}Defocus r=8(D8)\hspace{-0.7em}~& 53.34\% & 60.97\% & 51.40\% & 22.52\%\\\hline
      \hspace{-0.7em}Horiz. length=4\hspace{-0.7em}~& 75.37\% & 89.11\% & 90.04\% & 86.59\%\\
      \hspace{-0.7em}Horiz. length=8\hspace{-0.7em}~& 70.73\% & 83.32\% & 82.77\% & 71.50\%\\\hline
      \hspace{-0.7em}Vert. length=4\hspace{-0.7em}~& 75.57\% & 88.79\% & 89.84\% & 86.19\%\\
      \hspace{-0.7em}Vert. length=8\hspace{-0.7em}~& 70.75\% & 82.51\% & 81.86\% & 69.20\%\\\hline
      Camera~Shake & 58.91\% & 69.02\% & 66.66\% & 44.77\%\\\hline
      Gaussian~$\sigma=4$ & 56.34\% & 62.38\% & 49.80\% & 19.76\%\\
      Gaussian~$\sigma=8$ & 30.15\% & 17.39\% & 11.37\% & 3.41\%\\\hline
    \end{tabular}}\\~
}

\caption{Top-5 Accuracy on sharp and blurred versions of Imagenet Val images, using the original \textbf{VGG-16} network that was trained on sharp images. We report performance  for applying the network at different scales, where the input images were resized to make their smaller side fit the indicated scale value.}
\label{tab:origAll}
\end{table}

We use the original network model provided by the authors of \cite{vgg16}, who reported best performance by applying the network at two scales---256 and 512----where at each scale, the network (which has a receptive field of $224 \times 224$) was applied in a fully-convolutional way on all crops (with the network's natural stride of 32). The log-probabilities from all crops from all scales were then averaged. We adopt a similar methodology, but also apply the network at different individual scales---128, 256, and 512---as well as the suggested combination of 256 and 512. To apply the network at the scale of 128, which is smaller than the network's receptive field, we zero-pad the \emph{pool5} activations before passing them to the first fully connected layer.

Table \ref{tab:origAll} shows a clear drop in performance with increasing degrees of blur, for all types of blur kernels. Interestingly, this drop is steeper for bigger kernels in cases when the larger scale of 512 is included---by itself, or in combination with 256. Re-scaling to a larger size also increases the effective blur kernel size in the input to the network, and these results suggest that the benefits of robustness to object scale variation---the original motivation for multi-scale evaluation---are outweighed by the increasing effect of blur for larger kernels. However, going to the lowest scale of 128  hurts performance (except for the largest kernel considered in our evaluation: Gaussian $\sigma=8$), likely because it causes a mis-match between object sizes seen during training (which was done with scales 256 and 512).

Figure~\ref{fig:teaser} shows an example of the performance drop between sharp and blurred (with an $r=8$ defocus kernel) versions of an image. While the true  class was correctly identified as most likely with high probability in the sharp image, it is relegated to sixth place in the blurred version. Note however that the class probabilities for even the most likely class are much lower, and the distribution is closer to uniform with high entropy. This phenomenon is not specific to this example---in Table~\ref{tab:entropy}, we report the average entropy (and cross-entropy) values of the predicted class distributions across the entire validation set, for sharp images, and two levels of defocus blur. Looking at the ``Original'' column of the table (which also includes results from fine-tuning the network discussed later in Sec.~\ref{sec:finetune}), we see that the drop in accuracy in Table.~\ref{tab:origAll} is consistently accompanied by a significant increase in entropy. Therefore, while the model makes inaccurate predictions when used on unexpectedly low-quality images (compared to its training set), it does so with low-confidence.

\begin{table}[!t]
\centering{\footnotesize\renewcommand{\arraystretch}{1.0}
    \begin{tabular}{|c||c|c||c|c|}
      \hline
      &\multicolumn{2}{|c|}{Entropy}&\multicolumn{2}{|c|}{Cross-Entropy}\\
      \cline{2-3}\cline{4-5}
      &Original&Fine-tuned&Original&Fine-tuned\\\hline\hline
      Sharp & 1.1837 & 1.1215 & 1.2060 & 1.1449 \\
      Blur D4 & 2.6131 & 1.3629 & 2.6513 & 1.3882 \\
      Blur D8 & 3.8021 & 1.6553 & 3.8654 & 1.6819\\\hline
    \end{tabular}\\~}
  \caption{Entropy and cross-entropy of predicted class-distributions on sharp Imagenet Val images and those blurred with defocus blur (of radius 4 and 8). Results are from applying (at scale 256) the original VGG-16 network, and the version fine-tuned with a mix of sharp and blurry (D2,4,6,8) images.}
  \label{tab:entropy}
\end{table}

\begin{table*}[!t]
  \centering
  {\footnotesize\renewcommand{\arraystretch}{1.2}
    ~~~~~~~~~~~~~~~~~~~~~~~~~~~~~~~~~~~~~~~~~~~~~~~~{\bf Network Model Fine-tuned on:}\vspace{0.5em}\\%
\parbox[c]{1.5em}{\rotatebox{90}{\bf Test Blur Type~~~~~~~~~~}}\parbox[c]{41em}{
    \begin{tabular}{|c||c||c|c|c|c||c|}
      \hline
      & Original & \parbox[c]{2.5em}{Sharp\\\& D4}&\parbox[c]{6em}{\centering Sharp \&\\D2,D4,D6,D8} &\parbox[c]{2.5em}{Sharp\\\& D8}& Only D8 & \parbox[c]{3.7em}{\vspace{0.25em}\centering Sharp \&\\Camera\\Shake\vspace{0.25em}}\\\hline\hline
      Sharp & 90.88\% & 91.36\% & 90.59\% & 91.03\% & 16.10\% & 90.56\%\\\hline
      Defocus r=4 (D4) & 81.48\% & 89.84\% & 89.44\% & 85.39\% & 67.03\% & 87.79\%\\\hline
      Defocus r=8 (D8) & 60.97\% & 66.90\% & 87.01\% & 87.88\% & 88.43\% & 82.36\%\\\hline\hline
      Camera Shake & 69.02\% & 73.18\% & 80.78\% & 79.60\% & 60.52\% & 88.84\%\\\hline\hline
      Horiz. length=8 & 83.32\% & 81.09\% & 80.66\% & 80.62\% & 28.02\% & 88.52\%\\\hline
    \end{tabular}}\\~
}
\caption{Fine-tuning and Generalization. We report Top-5 accuracy on different versions of the Imagenet validation set blurred with different kernels, using network models fine-tuned with different distribution of blur kernels. Both training and evaluation are at scale 256. Fine-tuning with a mix of sharp and blurred images significantly improves performance of the network on blurred inputs at test time, with negligible penalty on sharp images. There is also a fair amount of generalization across defocus and camera shake blur, with fine-tuning on one improving performance on the other. However fine-tuning with defocus blurs slightly degrades performance for motion-blurred images, suggesting no generalization across blur types. We see a slight improvement in performance on images blurred with D8 by training only with that kernel (over a more diverse fine-tuning set), but this severely degrades performance for other blur types.}
\label{tab:ftune}
\end{table*}
\section{Robustness to Blur with Fine-tuning}
\label{sec:finetune}

Next, we investigate how much of the performance degradation seen in the previous section is simply due to a lack of blurred image samples in the training data. Other plausible explanations for this degradation include the possibility of a deficiency in the network architectures themselves (and not just their parameter values), or that blurred images simply lack the required semantically discriminative information needed to make accurate predictions. We largely rule out these possibilities by showing that fine-tuning the original VGG-16 for a small number of epochs on different combinations of sharp and blurred images nearly closes the performance gap caused by blur. We also explore the extent to which recognition can be made robust to variations in the type and degree of blur, by including different sets of blur kernels during fine-tuning.

We begin with the weights in the pre-trained model, and then fine-tune with two epochs at a learning rate of $10^{-3}$ followed by one epoch at $10^{-4}$, with a momentum value of $0.9$ and a batch-size of 128. We construct our training batches from the official Imagenet-2012 training set, shuffling the set at the beginning of each epoch. We  blur each image in the batch with a different selected kernel (or do not blur it at all) based on the chosen blur distribution. We resize the blurred image to either a fixed scale or a randomly selected scale, and take a random $224\times 224$ crop.

The original model was trained at multiple scales in order to build some degree of invariance to object scale~\cite{vgg16}. However, as described in the previous section, image re-scaling interacts with blur. Therefore, most of our experiments involve re-scaling the blurred training images to a fixed scale, with only one setting involving multiple scales. We try to build object scale invariance, by considering different scales \emph{prior} to the application of blur. Remember in our testing regime, a specific blur kernel is defined with respect to a fixed scale (of 384). So, we randomly re-scale training images to values around that scale (356, 384, and 410 in our experiments), and then apply blur. We then re-scale this blurred image to the target scale for the network by a factor equal to the ratio between the target training scale and the canonical blur scale (\ie, 384). This achieves variation in object scale, while ensuring that in all training examples that were blurred with the same selected kernel and fed to the network at the same post-blur scale, the effective size of the blur kernel is consistent.

\subsection{Fine-tuning with Different Blur Distributions}
\label{sec:bdist}

We begin by fixing the network scale at 256 (which yielded the best overall performance in Table \ref{tab:origAll}), and run multiple fine-tuning experiments with different combinations of sharp and blurred images. We consider different sets of defocus blur kernels, as well as camera shake kernels. For camera shake, we fine tune with a mix of sharp images (20\%) and those blurred with one of 10k camera shake kernels that are separate from, but generated with the same parameters as, those used for validation.

Table~\ref{tab:ftune} reports the performance of these different fine-tuned networks on different versions of the Imagenet validation set---including the original sharp images, and versions blurred by different blur kernels. We find that after fine-tuning with a diverse range of defocus blurs---specifically, a uniform mix of sharp images and those blurred with defocus blur of radii 2,4,6, and 8---the VGG-16 network begins to perform significantly better on images degraded by both defocus and camera-shake blur, \eg, increasing Top-5 accuracy on the radius-8 defocus blurred images from $61\%$ (for the original model) to $88\%$. Moreover, this comes at negligible cost to performance on sharp images, for which accuracy drops by less than a third of a percentage point. Indeed, fine-tuning with a mix of sharp images and only moderate (radius 4) defocus blur even leads to a slight improvement in sharp image accuracy, suggesting that moderate blurring can act as a form of data augmentation.

We see similar results when fine-tuning and testing on camera-shake blur. Interestingly, there appears to be a fair amount of generalization between blur types, with fine-tuning on defocus and camera-shake kernels significantly improving performance on the other. We also see generalization across degrees of defocus blur. Including examples of radius-4 blur along with sharp images also improves performance on radius-8 blur, and vice-versa. Fine-tuning on camera shake kernels improves performance on horizontal linear motion blur. However, fine-tuning on defocus kernels (and sharp images) alone seems to lead to a slight drop in accuracy on uniform motion blur. This is likely because images blurred with oriented motion kernels present significantly different statistics than defocus blur, with high-frequency attenuation being only along one direction in the former, and radially uniform in the latter.

We also see that if one has a-priori knowledge of the blur present in the input, there is some marginal improvement from using a network trained with a larger fraction of examples of that blur. However, the improvement is lower than the drop in performance for other blurs dropped from training. The most extreme example of this is when fine-tuning with all images blurred by the radius-8 defocus kernel, which severely degrades performance on sharp images and all other kinds of blur kernels.

\begin{figure}[!t]
\centering\includegraphics[width=\columnwidth]{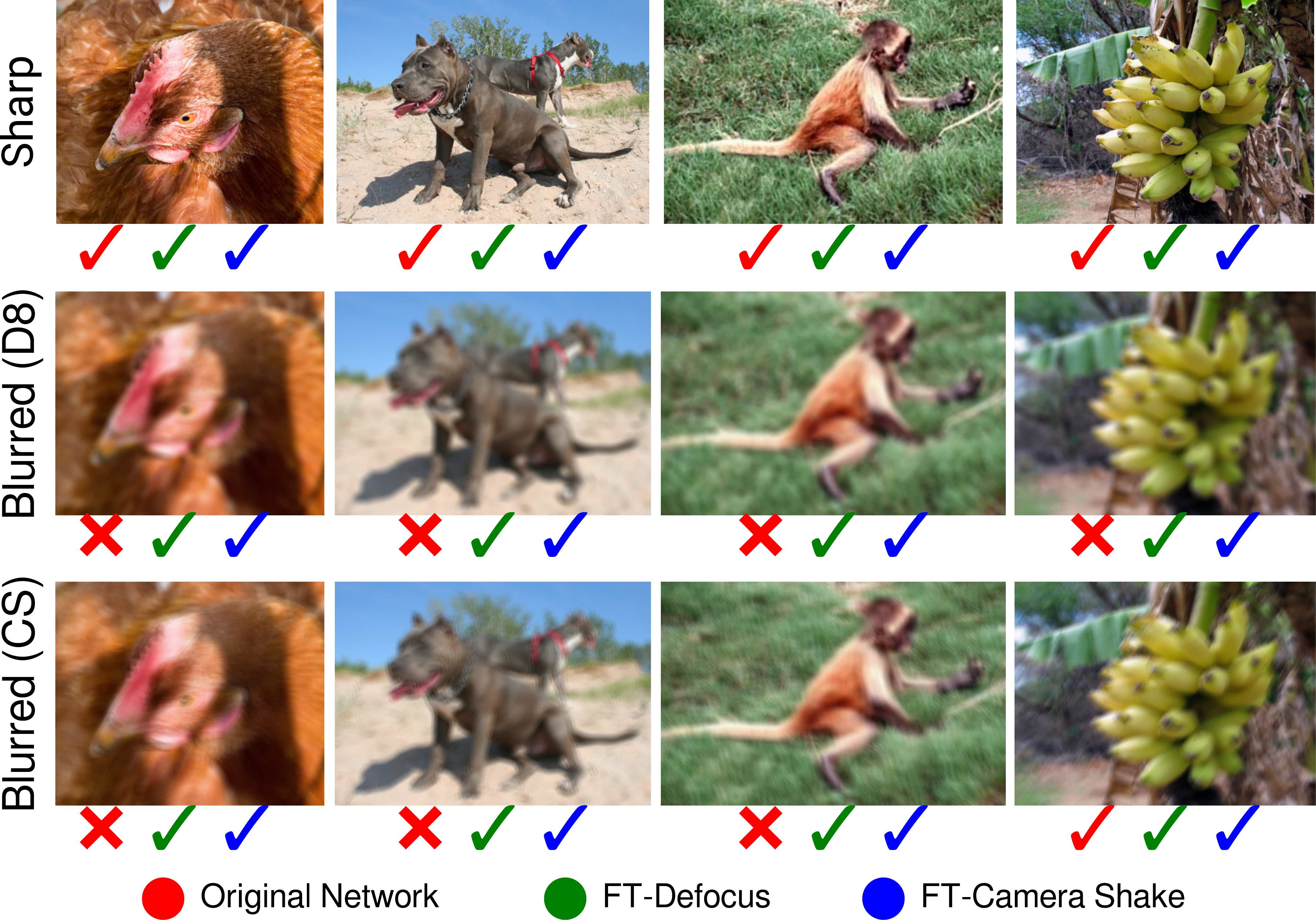}
\caption{Recognition on example blurred images. For each image, we indicate whether it was correctly classified (Top-5) by the original VGG-16 model (red), and ones fine-tuned on defocus (green) and camera shake (blue) kernels.}
\label{fig:eximg}
\end{figure}
Therefore, training with a diverse set of possible blurs appears to be the right strategy (at least at our chosen scale and range of blurs), unless one has an accurate estimate of the blur affecting the input. Figure~\ref{fig:teaser} includes the updated predicted class-distributions for our sharp and blurred example-pair from the model fine-tuned on the uniform defocus blur set. We see the network now identifies the correct class in both the sharp and blurred version of the image, and indeed produces very similar class distributions for both. In addition to identifying the correct class in the blurred image, the network makes this prediction with higher confidence. We report the average entropy of the class distributions from the fine-tuned model in Table~\ref{tab:entropy}. Fine-tuning not only improves the network's accuracy on blurred images, but also systematically increases its confidence in these predictions. Figure~\ref{fig:eximg} includes more examples of sharp and blurred images, and the ability of different network models to correctly determine their class labels.

\subsection{Fine-tuning for Different Scales}
\label{sec:ftscale}

Our decision to use a single scale of 256 to conduct the above fine-tuning experiments was based on the poorer performance of the original VGG-16 model at larger scales. However, a natural question is whether this trend continues to hold after fine-tuning. Therefore, we also fine-tune the model (beginning with the original weights) at a scale of 512, keeping the blur distribution fixed to the uniform setting (\ie, with a mix of sharp images and blurred with kernels of radii 2-8). We then evaluate this model on the validation set, applying it at the 512 scale, and compare its performance to the 256-scale version in Table~\ref{tab:scale}.

While fine-tuning also significantly improves accuracy on blurred images at the 512 scale in comparison to the original model, these values are measurably lower than those of the model fine-tuned and applied at scale 256. Also, this improvement comes at a greater cost to performance on sharp images, with accuracy dropping by nearly three percentage points compared to the original model. Thus, the relative advantage of smaller scales persists even after fine-tuning.

Next, we consider applying the network at both the 256 and 512 scales---this is the setting that achieves the highest accuracy on the sharp images with the original model. We test two versions of this approach. In the first, we fine-tune a single model on images randomly resized to one of these two scales (as was done for training the original model). As shown in Table~\ref{tab:scale}, this model does better than the 512 scale model on both sharp blurred images, and slightly worse than the 256 scale model on blurred images.

In the second case, we apply separate networks at each scale for classification---using the models fine-tuned at only their respective scales. This approach finally catches up to the 256 scale fine-tuned model on blurred images, and actually outperforms it by half a percentage point on sharp images. This is likely because for sharp images, both scales participate in classification, while on blurred images, the 512-scale model outputs higher-entropy distributions causing the average to revert to the output of the 256-scale model. While the per-scale fine-tuning strategy does have slightly better overall performance (when considering both sharp and blurred images), this marginal improvement is unlikely to justify the additional memory requirement of storing two versions of the model for most applications.
\begin{table}[!t]
  \centering{\footnotesize\renewcommand{\arraystretch}{1.0}
    \begin{tabular}{|c|c||c|c|c|}
      \hline
      Scale&Network&Sharp&D4&D8\\\hline\hline
      256 & Original & 90.88\% & 81.48\% & 60.97\%\\
      \cline{2-5}
          & Fine-tuned & 90.59\% & 89.44\% & 87.01\%\\\hline\hline
      512 & Original & 90.76\% & 66.86\% & 22.52\%\\
      \cline{2-5}
          & Fine-tuned & 87.86\% & 85.99\% & 81.54\%\\\hline\hline
      256+512 & Original & 92.17\% & 80.93\% & 51.40\%\\
      \cline{2-5}
           & \parbox[c]{6.6em}{\vspace{0.25em}\centering Fine-tuned\\(single network)\vspace{0.25em}}
           & 90.84\% & 89.17\% & 85.86\%\\
      \cline{2-5}
           & \parbox[c]{6.6em}{\vspace{0.25em}\centering Fine-tuned\\(per-scale)\vspace{0.25em}}
           & 91.10\% & 89.80\% & 87.09\%\\\hline
    \end{tabular}\\~}
  \caption{Top-5 Accuracies on sharp and blurry (defocus blur of radius 4 \& 8) versions of Imagenet Val images, when applying the network at different scales. We show results from the original network, and versions that are fine-tuned (with mix of sharp and D2,4,6,8) at their respective scales. For the multi-scale 256+512 evaluation, we show results from a single network fine-tuned with both scales, and from applying a separate network at each scale.}
  \label{tab:scale}
\end{table}
\begin{figure*}[!t]
  \centering
  \includegraphics[width=0.95\textwidth]{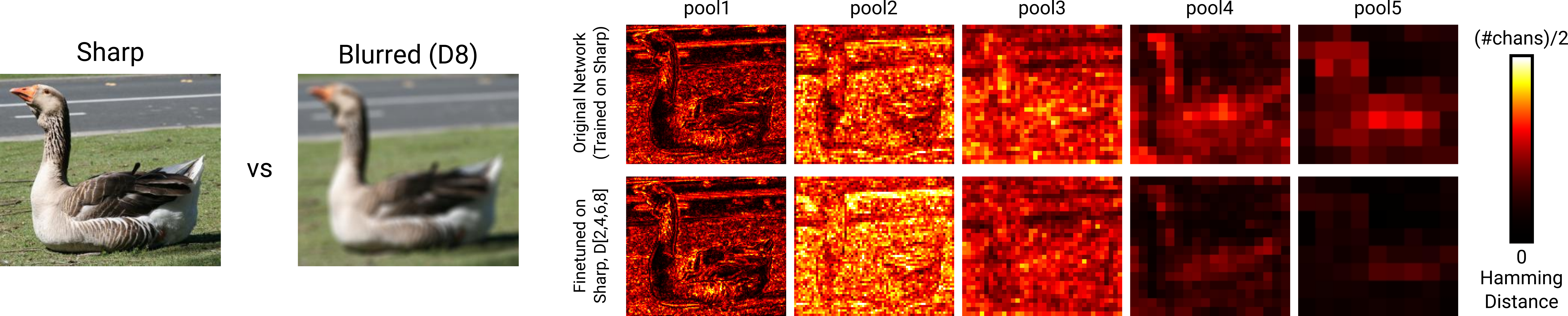}
  \caption{Disparity between corresponding layer activations on sharp and blurred versions of an example image, for different models at different scales. Each heat-map represents the Hamming distance between binarized feature vectors (\ie, if each channel is positive or zero) at corresponding locations in the sharp and blurred inputs. We visualize these distance maps for the different pooling layers in the VGG-16 architecture, rescaling the maps for all layers to be the same size, and normalizing them by the number of feature channels. We see that the model fine-tuned (at scale 256) on a mix of sharp and blurred images produces feature activations at higher layers that are relatively invariant to the presence of blur in the input image. }
  \label{fig:actcmp}
\end{figure*}

\subsection{Blur Invariance in Network Features}

The reliable performance of the fine-tuned model on a wide range of blur sizes suggests that the network learns to be invariant to blur. Does the network achieve this invariance early on, at the level of low-level image features in its initial layers, or at a more semantic level later in the network? We attempt to answer this question by looking at the similarity in activations of different layers of both the original and fine-tuned VGG models, when they are applied on the sharp and blurred versions of the same image. 

Specifically, we consider the feature maps at the output of the five pooling layers in the VGG-16 network. We convert the feature vector at every location into a binary string representing whether the each feature channel had a positive or zero response. In Fig.~\ref{fig:actcmp}, we visualize Hamming distances between corresponding binary strings produced from a sharp and blurred (by radius-8 defocus blur) versions of the same example image (at scale 256), for the original VGG-16 model and the model fine-tuned on a mix of sharp and blurred (defocus radius 2-8) images. We find that the original model produces different activations on the sharp and blurred inputs, at all layers. In contrast, the fine-tuned model is able to achieve a reasonable amount of blur invariance, with low distances between sharp and blurred activations at the fifth (and to a lesser extent the fourth) pooling layer. Note that the disparity in activations in the initial layers of the network remains high. We believe this may be because the initial layers have too small a receptive field to be able to reason about blur, or because blur invariance is easier to achieve at a higher semantic level.

\subsection{Comparison to Explicit Deblurring}

An alternative strategy to deal with blurred images is to explicitly deblur the input prior to recognition. This involves two steps---estimating the blur kernel, and deconvolution. Since kernel estimation is challenging, we measure the best-case performance of this strategy by applying a state-of-the-art deconvolution method~\cite{zoran} with the true blur kernel. Table~\ref{tab:deblur} reports results for using the original VGG-16 model on sharp, blurred, and then deblurred images, for the D8 defocus kernel---on a subset of the validation images. We see that deblurring (with knowledge of the blur kernel) also improves recognition accuracy, allowing the use of models trained only on sharp images. However, this comes at significant computational expense---\cite{zoran} takes roughly 4 minutes per images on a modern GPU.

Table~\ref{tab:deblur} also reports the performance of fine-tuned models applied directly on blurred images, on the same validation subset. We consider both the model fine-tuned on only on the D8 kernel---which corresponds to knowing that the blur at test time will be D8, and matches the knowledge in the deblurring results---as well as the one fine-tuned on a mix of defocus blurs (sharp \& D2,4,6,8)---which is more appropriate when the input blur isn't known. We find that incorporating robustness to blur directly in the recognition model performs as well or better, while avoiding the computational expense of deconvolution.
\begin{table}[!t]
  \centering{\footnotesize\renewcommand{\arraystretch}{1.0}
    \begin{tabular}{|c|c|c|c|c|}
      \hline
      Original&\hspace{-0.6em}Original on\hspace{-0.6em}~&\hspace{-0.6em}Original on\hspace{-0.6em}~&\hspace{-0.6em}FT-Mix on\hspace{-0.6em}~&\hspace{-0.6em}FT-D8 on\hspace{-0.6em}~\\
      on Sharp&\hspace{-0.6em}Blurred (D8)\hspace{-0.6em}~&\hspace{-0.6em}~Deblurred\hspace{-0.6em}~&\hspace{-0.6em}Blurred (D8)\hspace{-0.6em}~&\hspace{-0.6em}Blurred (D8)\hspace{-0.6em}~\\
\hline
      90.60\% &  61.57\% & 87.00\% & {\bf 87.15\%} & {\bf 88.3\%}\\
\hline
    \end{tabular}\\~}
    \caption{Top-5 accuracies on a subset (10 images per class) of validation images. We report performance of the original network (applied at scale 256) on sharp images, those blurred with D8, and then deblurred using \cite{zoran} with knowledge of the blur kernel. We compare the latter to applying the fine-tuned models directly on the blurred images, for versions fine-tuned with a mix of sharp and range of defocus kernels (FT-Mix), and on just D8 (FT-D8).}
  \label{tab:deblur}
\end{table}

\begin{figure*}[!t]
  \centering
  \includegraphics[width=0.77\textwidth]{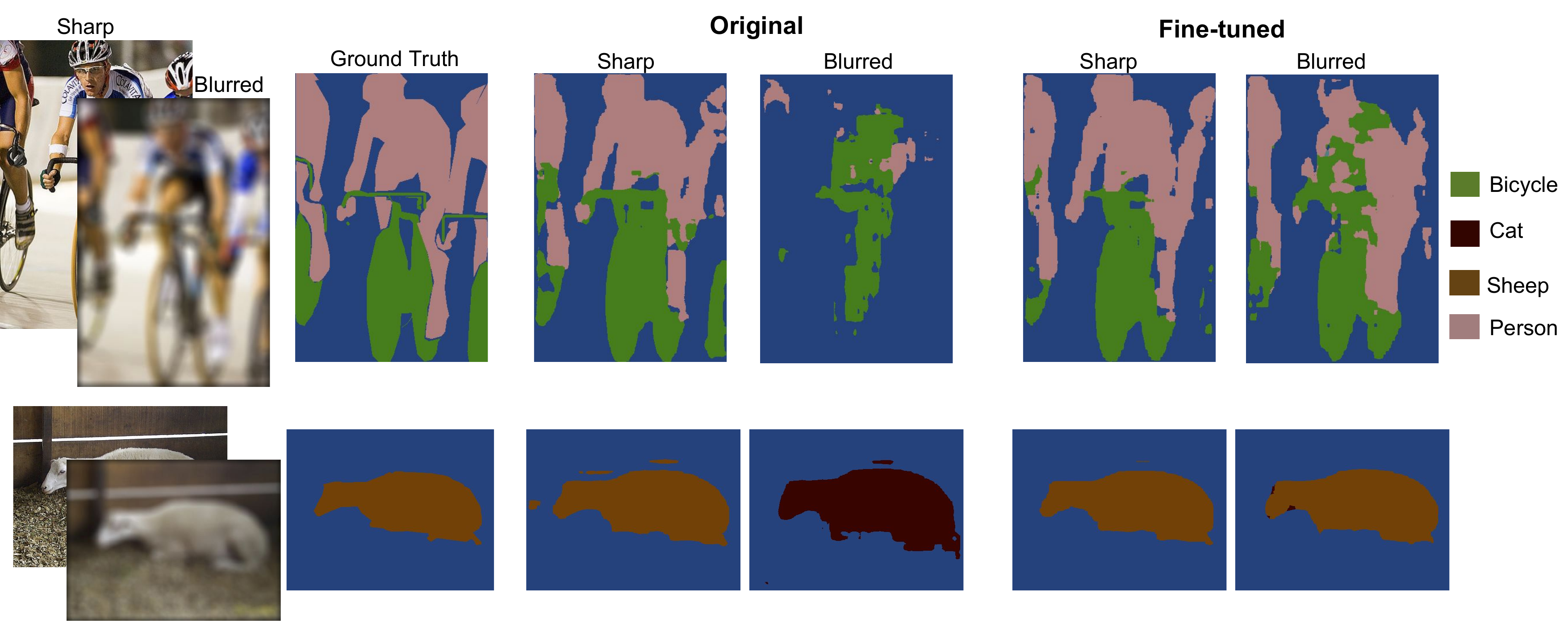}
  \caption{Semantic segmentation results on sharp and blurred images using the Zoomout~\cite{zoomout} network. We show results from the original model trained on sharp images, and one fine-tuned on a mix of sharp and blurred examples. Blur causes both errors in localization (top-row) and class identification (bottom-row) in the original model, but these errors are considerably reduced by fine-tuning.}
  \label{fig:voc}
\end{figure*}

\section{Semantic Segmentation on Blurred Images}

Next, we examine the effect of blur on convolutional neural network-based approaches to category-level semantic segmentation. Semantic segmentation is an especially interesting task in this context since it relies more on low-level image detail---that is attenuated by blur---to localize objects precisely at the pixel level.

We perform our experiments using the method from~\cite{zoomout} on the VOC2012 benchmark~\cite{voc}. This method learns a per-pixel classifier for object class on top of a feature-vector constructed from concatenated activations from multiple layers of VGG-16. This classifier is trained in conjunction with fine-tuning of the VGG-16 layers (the network is first pre-trained on ImageNet classification). We begin by evaluating the effect of blur on model weights made available by the authors of~\cite{zoomout}, that was trained on (largely sharp) images in the VOC training set and from the COCO segmentation database~\cite{coco}. We only consider different degrees of defocus blurs in this case, and use a similar methodology as before for generating blurred versions of the VOC2012 validation images, \ie, after resizing to a fixed scale of 384. However, since \cite{zoomout} uses anisotropic scaling to create the input (distorting the aspect-ratio) to its network, we perform scaling by matching the geometric mean, instead of the minimum, of width and height to 384.

Table~\ref{tab:voc} reports values of the standard mean Intersection-over-Union (mIOU) metric on sharp and blurred versions of the VOC validation set. In order to separately evaluate the degradation in boundary localization, we also report a version of this metric computed only at locations within a four pixel distance of class boundaries. Like for classification, we note that the pre-trained model suffers significant degradation in the presence of blur. However, this degradation appears to affect both general accuracy and localization. We see examples of this in Fig.~\ref{fig:voc} where in one instance, the model is able to identify object classes correctly in the blurred image, but performs a poorer segmentation between the classes and the background than in the sharp version. In the other example, the network produces a fairly high-quality segmentation of the foreground from the blurred image, but mis-identifies its class.

Next, we evaluate the effect of fine-tuning the model with a mix of sharp and blurred images. We fine-tune only on the VOC training set, for six epochs at a learning rate of $10^{-4}$, and two at a rate of $10^{-5}$. The performance of this fine-tuned model is also reported in Table.~\ref{tab:voc}. Like for image classification, we find that performance on blurred images improves with fine-tuning. Moreover, this improvement is seen in both general accuracy, as well as in regions close to boundaries. In the first example in Fig.~\ref{fig:voc}, we see that the fine-tuned network improves its ability to localize class boundaries on the blurred input, and is able to switch to the correct class label in the second input.

However, in contrast to classification, we find that the gap between sharp and blurred image performance remains larger after fine-tuning, and the improvements at a greater cost to sharp image performance. We believe this is due to the fact that this task is fundamentally more affected by blur. It requires identifying and separating objects, and the smearing of intensities across object boundaries due to blur makes this separation harder.

\begin{table}[!t]
  \centering{\footnotesize\renewcommand{\arraystretch}{1.0}
    \begin{tabular}{|c||c|c|c|c|}
      \hline
      &\multicolumn{2}{|c|}{Original}&\multicolumn{2}{|c|}{Fine-tuned}\\
      \cline{2-3}\cline{4-5}
      &All & Boundaries & All & Boundaries\\\hline\hline
		Sharp     &     70.0\%  & 55.5\% & 68.2\% & 54.2\%    \\\hline
		Blur D2   &     63.8\%  & 47.0\% & 64.8\% & 50.2\%   \\
		Blur D4   &     50.0\%  & 33.5\% & 58.7\% & 44.2\%   \\
		Blur D6   &     35.2\%  & 21.8\% & 51.3\% & 37.3\%    \\
		Blur D8   &     23.1\%  & 13.2\% & 43.0\% & 30.5\%    \\\hline
    \end{tabular}\\~}
  \caption{mIOU accuracy on sharp and blurred versions of the VOC val images, using the Zoomout network with original and fine-tuned weights. In addition to average accuracy over all pixels, we also separately calculate mIOU over pixels that within a four pixel neighborhood of class boundaries.}
  \label{tab:voc}
\end{table}

\section{Conclusion}

State-of-the-art network models trained on high-quality image datasets make unreliable, albeit low-confidence, predictions when they encounter blur in their inputs. In this work, we found that much of this unreliability is due to an inability to generalize from their sharp training sets, and that fine-tuning these models for a relatively small number of epochs with blurred training examples significantly improves their performance on blurry inputs. Moreover, we showed that standard architectures are able to deal with a diverse range of blurs, by learning to produce internal representations that are invariant to blur.

Our analysis provides insights for building and deploying vision systems in real-world settings where blur may be present. More broadly, we expect our findings to be relevant for other forms of imaging non-idealities beyond blur. In future work, we plan to explore un-supervised ways of achieving robustness to image artifacts---\eg, when we have access to examples of distorted natural images, but not to a precise model for the distortion.

{\small
\paragraph{Acknowledgments} This work was partially supported by the NSF under award no.~RI:1409837.
\bibliographystyle{ieee}

}
\clearpage
\onecolumn

\section*{Supplementary Results on ResNet-51}
We also report classification results on blurred images with the more recent ResNet~\cite{resnet} architecture in Table~\ref{tab:resnet}, considering various levels of defocus blur. We use the pre-trained model weights provided by the authors, and apply the network on five crops with two different scale settings: 256, and a combination of 256 and 512.  We find that while the absolute accuracies of the ResNet model are slightly higher---for both sharp and blurred images---than for VGG-16, there is a similar drop in accuracy with increasing blur. This confirms that the drop in accuracy due to blur is not peculiar to the VGG-16 architecture.

\renewcommand{\thetable}{A}
\begin{table}[!h]
  \centering{\footnotesize\renewcommand{\arraystretch}{1.0}
    \begin{tabular}{|c||c|c|}
      \hline
      \bf Scale&$256$&$256+512$\\\hline\hline
      Sharp   & 92.89\% & 93.44\% \\
      Blur D2 & 90.40\% & 90.82\% \\
      Blur D4 & 85.20\% & 83.54\% \\
      Blur D6 & 77.07\% & 69.52\% \\
      Blur D8 & 66.14\% & 55.94\% \\\hline
    \end{tabular}\\~}
  \caption{Top-5 Accuracies for the pre-trained \textbf{ResNet-51} model on sharp and blurred (with defocus blur of different radii) versions of Imagenet Val images. We observe a similar drop in performance with increasing blur as in Table ~\ref{tab:origAll}.}
  \label{tab:resnet}
\end{table}

\end{document}